\documentclass[letterpaper, 10 pt, conference]{ieeeconf}  

\IEEEoverridecommandlockouts                              

\usepackage{graphicx} 
\usepackage{amsmath} 
\usepackage{amssymb}  

\usepackage{comment} 
\usepackage[table,xcdraw]{xcolor}
\usepackage{cite}

\title{\LARGE \bf
Simultaneous Locomotion Mode Classification and Continuous Gait Phase Estimation for Transtibial Prostheses\\[-2ex]
}

\author{Ryan R. Posh$^{1}$, Shenggao Li$^{2}$, and Patrick M. Wensing$^{2}$
\thanks{*This work was funded by NSF grants DGE-1841556 \& CMMI-1943703.}
\thanks{$^{1}$R. Posh is now with the department of Robotics, University of Michigan, Ann Arbor, MI 48109, USA 
        {\tt\small poshr@umich.edu}}%
\thanks{$^{2}$S. Li and P. Wensing (and previously R. Posh) are with the Department of Aerospace and Mechanical Engineering, University of Notre Dame, Notre Dame, IN 46556, USA}
}

\usepackage{amsmath}
\usepackage{amssymb}
\usepackage{algorithm}
\usepackage{algorithmic}
\renewcommand{\Lsh}{{\rm{L,sh}}}
\renewcommand{\Rsh}{{\rm{R,sh}}}
\newcommand{\Lth}{{\rm{L,th}}}
\newcommand{\Rth}{{\rm{R,th}}}
\newcommand{\Lft}{{\rm{L,ft}}}
\newcommand{\Rft}{{\rm{R,ft}}}

\newcommand{\thetaType}[1]{\theta^{#1}}

\newcommand{\thetaLth}{\thetaType{\Lth}}
\newcommand{\thetaLsh}{\thetaType{\Lsh}}
\newcommand{\thetaLft}{\thetaType{\Lft}}
\newcommand{\thetaRth}{\thetaType{\Rth}}
\newcommand{\thetaRsh}{\thetaType{\Rsh}}
\newcommand{\thetaRft}{\thetaType{\Rft}}

\newcommand{\thetab}{\bar{\theta}}
\DeclareMathOperator{\argmin}{arg\,min}

\begin{document}

\maketitle
\thispagestyle{plain}
\pagestyle{plain}

\begin{abstract}

Recognizing and identifying human locomotion is a critical step to ensuring fluent control of wearable robots, such as transtibial prostheses. In particular, classifying the intended locomotion mode and estimating the gait phase are key. In this work, a novel, interpretable, and computationally efficient algorithm is presented for simultaneously predicting locomotion mode and gait phase. Using able-bodied (AB) and transtibial prosthesis (PR) data, seven locomotion modes are tested including slow, medium, and fast level walking (0.6, 0.8, and 1.0 m/s), ramp ascent/descent (5 degrees), and stair ascent/descent (20 cm height). Overall classification accuracy was 99.1$\%$ and 99.3$\%$ for the AB and PR conditions, respectively. The average gait phase error across all data was less than 4$\%$. Exploiting the structure of the data, computational efficiency reached 2.91 $\mu$s per time step. The time complexity of this algorithm scales as $O(N\cdot M)$ with the number of locomotion modes $M$ and samples per gait cycle $N$. This efficiency and high accuracy could accommodate a much larger set of locomotion modes ($\sim$ 700 on Open-Source Leg Prosthesis) to handle the wide range of activities pursued by individuals during daily living.
\end{abstract}

\section{INTRODUCTION}






Assistive devices like lower-limb prostheses and exoskeletons have the potential to enhance the quality of life for a wide range of individuals. A key challenge in realizing the full potential of these devices is to ensure that the control is consistent and fluent with the user's intent~\cite{hoffman2019evaluating}. These human-centered robots depend on accurate recognition and interpretation of the user's intended movements to ensure a positive user experience and to maximize the assistive benefit~\cite{gambon2020effects}. Recognizing and analyzing human gait can also be used clinically to identify healthy or pathological gait~\cite{hutabarat2020quantitative, caramia2018imu, carcreff2018best, filli2018profiling} or to assess rehabilitation progress~\cite{ilias2017using, del2015validation}. Therefore, identifying highly accurate ways to classify and monitor human locomotion is key to the advancement of rehabilitation robotics and assistive devices. 

\begin{figure}[b!]
    \centering
    \includegraphics[width = 0.85\linewidth, trim={1cm 12cm 0cm 0.3cm},clip]{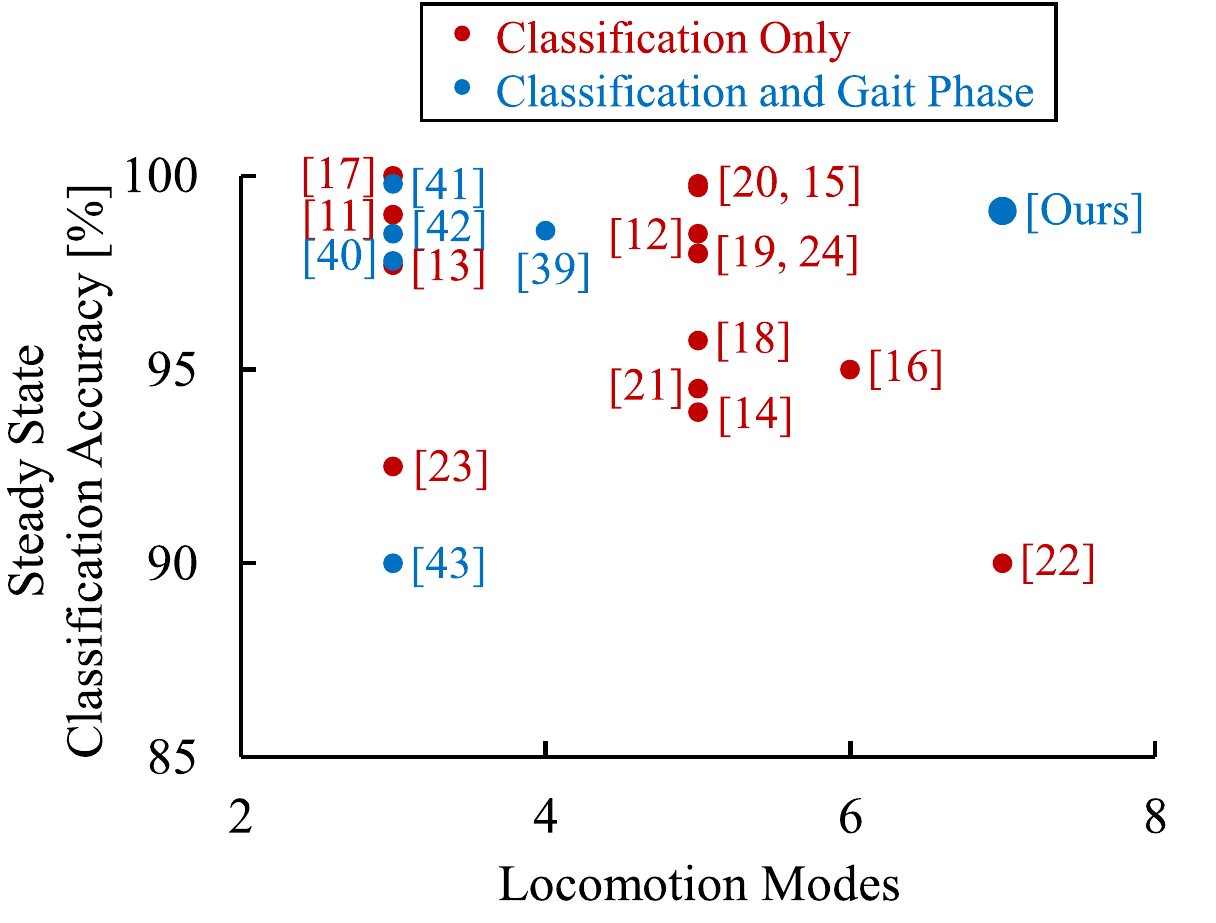}
    \caption{The accuracy and number of locomotion modes present in representative locomotion mode classification literature. Studies in red present classification only and studies in blue classify the locomotion mode and estimate the gait phase. } 
	\label{fig:Literature}
\end{figure}

\begin{figure*}[t!]
    \centering
    \includegraphics[width = 0.7\linewidth, trim={0cm 18.2cm 0cm 0cm},clip]{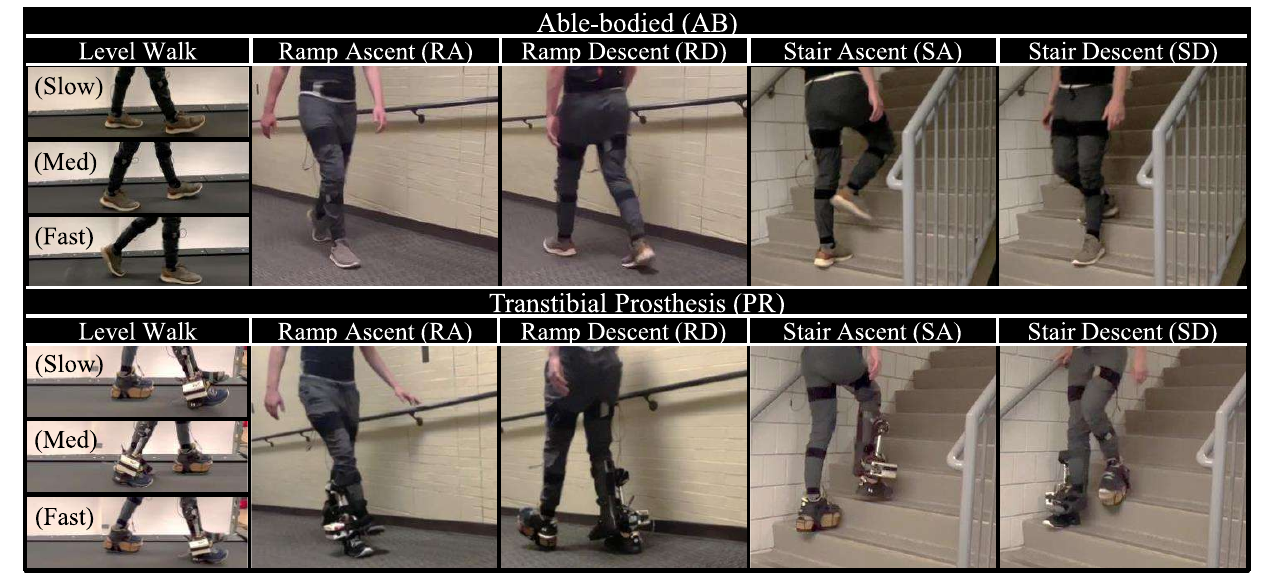}
    \caption{Experimental data was collected from 7 different locomotion modes: slow/medium/fast walking, ramp ascent/descent, and stair ascent/descent), with 2 different conditions: able-bodied (AB) and with the Open-Source Leg (OSL) transtibial prostheses (PR).} 
	\label{fig:Data_Collection}
    \vspace{-0.5cm}
\end{figure*}

Herein, the recognition and identification of human locomotion consists of two primary goals: classifying the intended locomotion mode (such as level walking, ramp walking, or stair climbing) and monitoring the progression of gait within each stride. Traditionally, for control of prostheses and exoskeletons, these goals are pursued consecutively in a hierarchical control framework. The highest level determines the user's activity intent, the middle level determines gait phase and translates that into device commands, and the lowest level implements those commands~\cite{tucker2015control}. 

High-level locomotion mode classification has primarily paired various wearable sensors with machine learning (ML) algorithms~\cite{labarriere2020machine, camargo2021machine}. Inertial measurement units (IMUs) are frequently employed in these frameworks as they can be accessible and relatively easily worn~\cite{gao2020imu, bartlett2017phase}. ML approaches such as linear discriminant analysis (LDA,~\cite{bartlett2017phase, young2013training, chen2014locomotion}), support vector machines (SVMs,~\cite{huang2011continuous}), Gaussian mixture models (GMMs,~\cite{varol2009multiclass, shin2021locomotion}), dynamic Bayesian networks (DBNs,~\cite{young2014intent, simon2016delaying}), convolutional neural networks (CNNs,~\cite{su2019cnn, lee2019classification, feng2019strain, le2024transfer}), and others have been used to distinguish between standard locomotion modes.

Once high-level classification is complete, the appropriate mid-level controller can be selected. To synchronize the mid-level commands with the user's gait, however, the gait stride progression must be identified. The gait phase can be defined as the progression from one heel strike to a consecutive same-side heel strike. Finite-state machines (FSMs) monitor gait phase via discrete phases, such as early stance, late stance, and swing~\cite{sup2007design, posh2023finite, vu2020review}. Gait cycle progression can also be continuously estimated by monitoring a phase variable. Phase variables are biomechanic markers that progress monotonically with gait phase, such as the foot center of pressure~\cite{gregg2013experimental, gregg2013towards} (during stance only), thigh kinematics~\cite{villarreal2014survey, cortino2023data}, and tibia kinematics~\cite{holgate2009novel, posh2023calibration, posh2024hybrid}. 
Gait phase has most commonly been measured during level-ground walking, but the wide range of achievable activities in the real world demands expanding these techniques to other activities~\cite{kang2021real}. 

Most approaches focus on either locomotion mode classification or on gait phase estimation~\cite{xu2021noninvasive}, while some have pursued both (\!\!\cite{martinez2017simultaneous, wu2019locomotion, zhang2022gait, weigand2022continuous, zhang2023real}, blue in Fig.~\ref{fig:Literature}). 
While these approaches demonstrate that IMUs on the leg segments can be an effective sensing strategy, two primary limitations persist. The first is that very few locomotion modes are often used for classification, and these modes are often substantially different from one another, such as standing and ascending stairs. While this simplification leads to higher classification accuracy, using a sparse set of locomotion modes is not representative of the nearly infinite task landscape of human locomotion. Furthermore, misclassifications that occur between dissimilar locomotion modes would lead to control actions that are significantly different from expected, potentially leading to falls~\cite{zhang2014effects}. 

The second perceived limitation of many ML strategies for classification and gait phase estimation is that 
the resulting predictions themselves lack transparency. While the model outputs may offer predictions, ML approaches typically do not provide insight into how or why those predictions were made. This insight could be critical information when analyzing misclassifications or when assessing user progress in a rehabilitation setting. Thus, in this study, we present a novel and interpretable approach to simultaneously identify the locomotion mode and gait phase, and implement it with able-bodied data and with a robotic transtibial prosthesis in a bypass configuration. Figure~\ref{fig:Data_Collection} shows these two conditions, along with the locomotion modes selected for identification, which are level treadmill walking (at 0.6, 0.8, and 1.0 m/s), fixed ramp ascent and descent (5-degree slope), and stair ascent and descent (20 cm stair height).

This approach is computationally efficient (2.91 $\mu$s per time step), which allows for a large number of reference locomotion modes with small differences from one another to be predicted with high accuracy (7 modes, $>99\%$). The gait phase estimation ($< 4\%$ average error) is done continuously to allow smooth continuous control of wearable robotics. The contributions herein include 1) the development of a novel, interpretable, computationally efficient algorithm to simultaneously predict locomotion mode and gait phase, 2) the assessment of this algorithm on healthy able-bodied gait and asymmetrical gait from a transtibial prosthesis, and 3) the demonstration of high steady-state classification accuracy and phase estimation while considering a large set of locomotion modes compared to the literature (see Fig.~\ref{fig:Literature}).










\section{GAIT and PHASE ESTIMATION}

%
%
%

%
%
%
%

%
%
%
%
%

\begin{figure*}[t!]
    \centering
    \includegraphics[width = 0.8\linewidth, trim={0cm 24cm 0cm 0cm},clip]{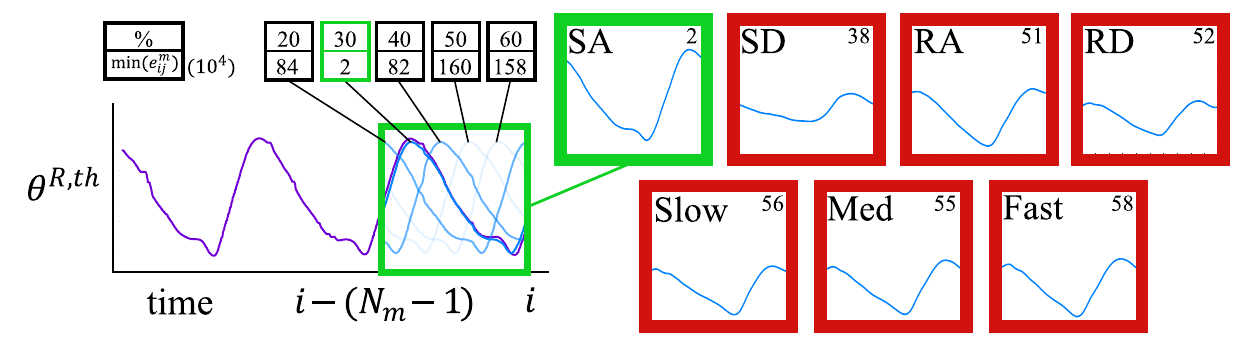}
    \caption{A window of historical data is compared to sliding locomotion mode kernels. Locomotion modes include slow/medium/fast walking (Slow, Med, Fast), ramp ascent/descent (RA, RD), and stair ascent/descent (SA, SD). The kernel with the lowest sum of squares error (SSE) is selected as the current locomotion mode (shown as SA, SSE = 2), and the sliding configuration that results in that minimum SSE determines the gait phase (shown as $\phi = 30\%$).} 
	\label{fig:Schematic}
    \vspace{-0.5cm}
\end{figure*}

%
%

\subsection{System Overview}

An overview of the system architecture is shown in Fig.~\ref{fig:Schematic}. The overall control framework takes as input the current left and right global thigh ($\thetaLth$ and $\thetaRth$), and shank ($\thetaLsh$ and $\thetaRsh$) angles, 
and outputs the predicted locomotion mode and the corresponding estimate of the gait phase progression for that mode. Using the training data, a baseline trajectory for each locomotion mode (referred to herein as a kernel) is identified. At each time instant thereafter, each kernel is compared to the most recent historical data, and the kernel with the best matching is selected as the predicted locomotion mode. The phase shift of that kernel that results in the best overall matching is used to estimate the gait phase. Note that this process considers kernel matching for all four segment angles, while Fig.~\ref{fig:Schematic} shows only $\thetaRth$.

\subsection{Steady-State Kernel Generation}

Steady-state kernels for each locomotion mode and each segment angle are generated offline using the training data (Section~\ref{sec:processing}). Once the data are segmented into strides, the average thigh and shank angle trajectories as a function of gait phase are stored as kernels. Unlike neural-network-based algorithms, this model-based training requires only a small number of steady-state strides ($\sim$ 10) from the individual and is computationally light (averaging). 
Seven kernels are generated representing the seven locomotion modes (three level walking speeds, ramp ascent/descent, and stair ascent/descent), with each kernel containing segment angle trajectories (bilateral thigh and shank).

\subsection{Mode and Phase Prediction}

For notation, let $k^{m} \in \mathbb{R}^{4 \times N_m}$ represent the individual kernel for locomotion mode $m$. 
Modes of 0.6 m/s level walking, 0.8 m/s level walking, 1.0 m/s level walking, ramp ascent, ramp descent, stair ascent, and stair descent are indicated by $m \in \{{\rm  Slow, Med, Fast, RA, RD, SA, SD}\}$, respectively. Each kernel $k_{m}$ has $N_{m}$ columns, where $N_{m}$ is the average number of data points during a stride of locomotion mode $m$ from the training data.
For example, at the approximately 230 Hz sampling rate, $N_{\rm Slow} = 392$, whereas $N_{\rm Fast} = 280$ in the AB condition,
as the stride duration is decreased with increased speed. Each column $j$ of the kernel matrix $k^m$ represents the average joint angles $\thetab_j^m$ observed at the $j$-th sample of the gait so that:
\[
k^m = \begin{bmatrix} \thetab_1^m  & \cdots & \thetab_{N_m}^m \end{bmatrix}
\]
Two copies of the kernel columns are then concatenated so that they can be compared to the most recent historical data in a circular sliding manner, giving the final kernel $K^m$ as 
\begin{equation}
    K^m = 
    \begin{bmatrix}
    k^{m}, & k^m
    \end{bmatrix} 
    \in \mathbb{R}^{4 \times 2N_{m}},
\end{equation}

We then compare the columns of this kernel matrix to data measured directly from the prosthesis. We construct a data vector $d_i = [\thetaRth_i, \thetaLth_i, \thetaRsh_i, \thetaLsh_i]^\top \in \mathbb{R}^{4}$ that contains the four segment angles for a given time instant $i$. 
To compare $K^m$ to the most recent historical data, we construct a data history matrix 
\begin{equation}
    D_i^m = 
    \begin{bmatrix} d_{i-(N_m-1)} & \cdots & d_i \end{bmatrix} 
    \in \mathbb{R}^{4 \times N_{m}},
\end{equation}
We then compare $D^m_i$ to sub-blocks of $K^m$ with a sliding window. Specifically, we let 
\[
K^m_j = K^m[:, j+1:j+N_m]
\]
denote the sub-block of the kernel matrix that ends with $\thetab_{j}^m$ (i.e., the average gait data for one cycle with $\thetab_j^m$ as the last column in the matrix). 
A complete match between $D_i^m$ and $K_j^m$ during the $i$-th timestep online would indicate that the user is at time-step $j$ of the average gait cycle for mode $m$. 

To assess all possible locations $j$ in in the gait mode, we consider the matrix of errors:
$E^m_{ij} = D^m_i - K^m_j$ 
and the sum of its squares:
\begin{equation}
e^{m}_{ij} = \| E^m_{ij}\|_{F}^2 
\end{equation}
for all $\forall j \in \{1, \cdots, N_m\}$ where $\|\cdot\|_{F}$ denotes the Frobeius norm, which takes the following form for any $A \in \mathbb{R}^{n \times m}$
\[
\|A\|_{F}^2 = \sum_{i=1}^n \sum_{j = 1}^m A_{ij}^2 = {\rm trace}(A^T A)\,.
\]


The best matching window between $D^m$ and $K^m$ is given by the index $j$ that minimizes $e_{i,j}^m$. This process is repeated for all locomotion modes ($m = \{ {\rm Slow, Med, Fast, RA, RD, SA, SD}\}$), and the locomotion mode with the lowest $e^m_{ij}$ is selected as the predicted locomotion mode $m^*$. 

Once the mode is classified, the index $j^* \in [1,N_m]$ that minimizes $e_{i,j}^{m^*}$ is normalized by $N_{m^*}$ to get the estimated gait phase $\phi = j^*/N_m \in (0,1]$. Therefore, the resolution at which each locomotion mode can estimate the gait phase is defined by $1/N_m$, allowing locomotion modes with longer stride durations to have finer resolution.
The finest detectable change in gait phase was for PR stair descent (0.23$\%$ of gait) and the coarsest was for AB fast walking (0.36$\%$ of gait).

\begin{figure}[b!]
    \centering
    \includegraphics[width = 0.8\linewidth, trim={0cm 13cm 10.6cm 0cm},clip]{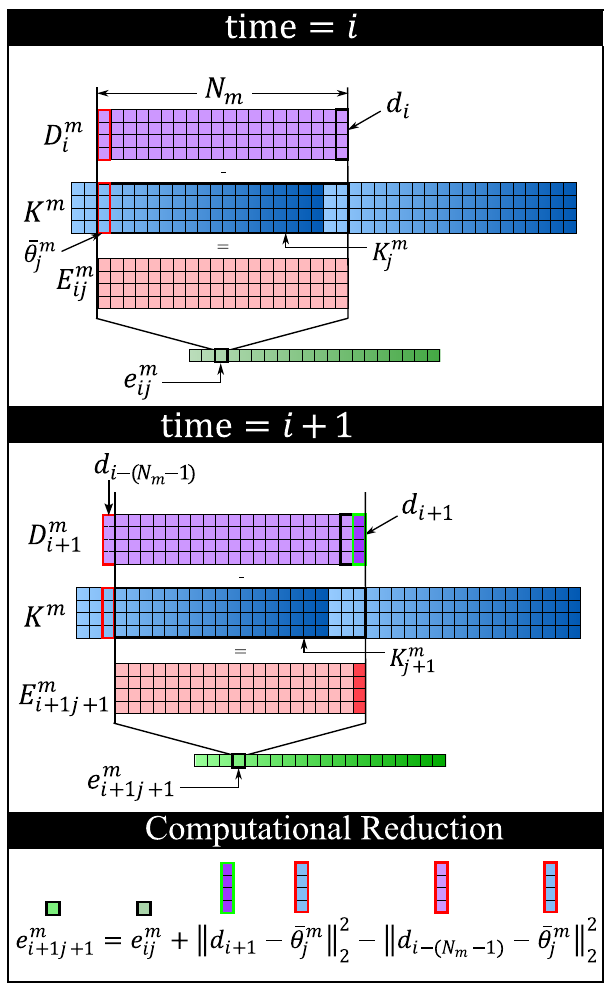}
    \caption{The sliding kernel matrix $K_m$ is compared to historical data $D_m$ via the sum of squares error (SSE). All calculations can be repeated for each time step, such as time = t (left) and time = t+1 (right), or computation can be significantly reduced by exploiting the many shared computations between subsequent time steps (bottom).} 
	\label{fig:Matrix}
    \vspace{-0.5cm}
\end{figure}

\subsection{Computational Reduction}

At each time instant, algorithm~\ref{alg:naive} calculates the errors $e_{ij}^m$ (each computationally $\mathcal{O}(N_m)$) for each $N_m$ sliding window configuration and repeats this for $M = 7$ locomotion modes. Therefore, the overall computational cost is $\mathcal{O}(N^2 \cdot M)$ where $N$ represents the max of all $N_m, m \in [1,M]$. This algorithm is computationally expensive at each time interval, and the computational cost scales linearly with the number of locomotion modes and quadratically as the kernel length increases. For implementation in real time, it is undesirable to reduce the number of locomotion modes that can be predicted or to decrease the data collection frequency. Alternatively, computational cost can also be reduced by exploiting the structure of the matrices used in the algorithm.



In particular, the vast majority of the individual error calculations are repeated in consecutive time steps due to much of the data history remaining the same between consecutive time steps, and all of the kernels remaining unchanged. 
In particular, note that the $j$-th element of the data-kernel comparison results $e_{ij}^m$ can be written as
\begin{equation}
e_{ij}^m = e_{i-1, j-1}^m + \|d_i - \thetab_{j-1}^m\|^2_2 - \|d_{i-N_m}-\thetab_{j-1}^m\|_2^2
\label{eq:theBigKahuna}
\end{equation}
which, intuitively, computes $e_{ij}^m$ by sliding both the kernel and the data back in time (to re-use $e_{i-1,j-1}^m$ from the previous timestep) while accounting for effects of the new data $\theta_i$ and removing those of the stale data $\theta_{i-N_m}$that has now existed the window of comparison. This process is depicted in Fig.~\ref{fig:Matrix}. By defining subscript $0$ as representing $N_m$, Eq.~\eqref{eq:theBigKahuna} still holds for $j=1$.

\begin{algorithm}[b]
\caption{Naive method for mode/phase computation}
\begin{algorithmic}[1]
\REQUIRE Kernel Matrices $K^m$
\REQUIRE Data history $\theta_i,\ldots, \theta_{i-N}$
\FOR{$m =1, \ldots, M$}
    \STATE $D^i_m = [\theta_{i-(N_m-1)},\, \ldots,\, \theta_i]$
    \FOR{$j = 1,\ldots, N_m$} 
        \STATE $K_j^m  = K^m[:, j+1:j+N_m]$
        \STATE $E_{ij}^m = D^m_i-K^m_j$
        \STATE $e_{ij}^m = \| E_{ij}^m\|_{F}^2$ \quad \tt{/* O(N) */}
    \ENDFOR
\ENDFOR
\RETURN $(m^*, j^*) = \argmin_{m, j} e_{ij}^m$
\end{algorithmic} \label{alg:naive}
\end{algorithm}

\begin{algorithm}[b]
\caption{Efficient method for mode/phase computation}
\begin{algorithmic}[1]
\REQUIRE Kernel Matrices $K^m$ \\[.25ex]
\REQUIRE Data history $\theta_i,\ldots, \theta_{i-N}$ \\[.25ex]
\REQUIRE Past errors $e_{i-1,j}^m$ \\[.25ex]
\FOR{$m =1,\ldots, M$}
    \FOR{$j = 1,\ldots, N_m$}
        \STATE $e_{ij}^m = e_{i-1, j-1}^m + \|\theta_i - \thetab_j^m\|^2_2 - S_{i-N_m,j}$
        \STATE $S_{i,j} = \|\theta_i - \thetab_j^m\|^2_2$ 
    \ENDFOR
\ENDFOR
\RETURN $(m^*, j^*) = \argmin_{m, j} e_{ij}^m$
\end{algorithmic} \label{alg:eff}
\end{algorithm}

Note that \eqref{eq:theBigKahuna} can be evaluated with constant computational complexity (i.e., carrying out a single one of these computations does not have demands that scale with $N$ or $M$). In this way, if the errors $e_{i-1,j}^m$ are available from the previous timestep, Algorithm~\ref{alg:naive}, which had computational cost of $\mathcal{O}(N^2 \cdot M)$, can now be replicated with only $\mathcal{O}(N \cdot M)$ operations (see Alg.~\ref{alg:eff}). 
Additionally, the squared term $\|\theta_{i-N_m}-\thetab_{j-1}^m\|_2^2$ computed $N_m$ steps earlier can be cached using $\mathcal{O}(N^2 \cdot M)$ memory. This caching allows for nearly half the computation time reduction and eliminates the storage of historical data.
In practice, the error array $e_{ij}^m$ and square term cache matrix $S$ are initialized to zeros. Therefore, the calculations from the reduction method are mathematically equivalent after $N_m$ time steps, at most about 1.85 seconds (one time period $T$ of SA in the PR condition) after data collection begins.

\section{EXPERIMENTS and RESULTS}
\subsection{Data Collection}

To evaluate the locomotion classification, gait phase estimation, and computational reduction approaches proposed herein, data was collected and processed for later offline analysis. Two datasets were collected; the first from a single able-bodied (AB) individual (180 cm height, 75 kg mass), and the second from that same individual walking on a unilateral transtibial robotic prostheses (PR), namely the version 1 Open-Source Leg (OSL)~\cite{azocar2018design} via a bypass adapter. When walking with the prosthesis, a passive controller was employed, defined by constant ankle impedance parameters of 0 degrees equilibrium angle, 0.09 Nm/deg/kg stiffness, and 0.075 Nms/deg/kg damping, intended to approximate the behavior of a passive ankle-foot prosthesis. A contralateral shoe lift was donned to accommodate leg length difference from the adapter.

For all conditions, the individual wore IMUs on the thighs and shank via the XSENS MVN Link motion capture suit for real-time measurement of the global thigh and shank angles, 
as well as the bilateral foot angles ($\thetaRft$ and $\thetaLft$). Only the four angles of the bilateral thigh and shank serve as inputs to the mode/phase prediction, while the bilateral foot angles were used strictly for heel strike (HS) identification (Sec.~\ref{sec:processing}). 
The system operated at approximately 230 Hz, bounded by the wireless XSENS communication frequency.

\subsection{Data Processing}
\label{sec:processing}

For both AB and PR conditions, two minutes of steady-state walking data were collected at each treadmill speed, 10 trials of ramp ascent/descent, and 5 trials of stair ascent/descent. The asymmetrical step-to-step gait was employed for PR stair descent, as is commonly recommended by clinicians~\cite{hafner2007evaluation}, while the step-over-step gait was used for PR stair ascent and AB stair ascent/descent, highlighting the ability of this approach to be used for a wide range of gait types. The data were then split into training data and evaluation data 
at a 1:1 ratio for level and ramp walking, and a 3:2 ratio for stair ascent/descent modes (due to 5 trials each).
In level-ground and ramp walking, right-side (or prosthesis-side) HS was identified by peaks in $\thetaRft$~\cite{das2019novel}, while peaks in $\thetaRsh$ were used during stair ascent and descent and manually verified. HS identification was only used to calibrate the training data and to serve as ground truth gait phase for the evaluation data.

\subsection{Activity Classification}

\begin{figure}[t!]
    \centering
    \includegraphics[width = 0.8\linewidth, trim={1.8cm 10.6cm 1.8cm 1.5cm},clip]{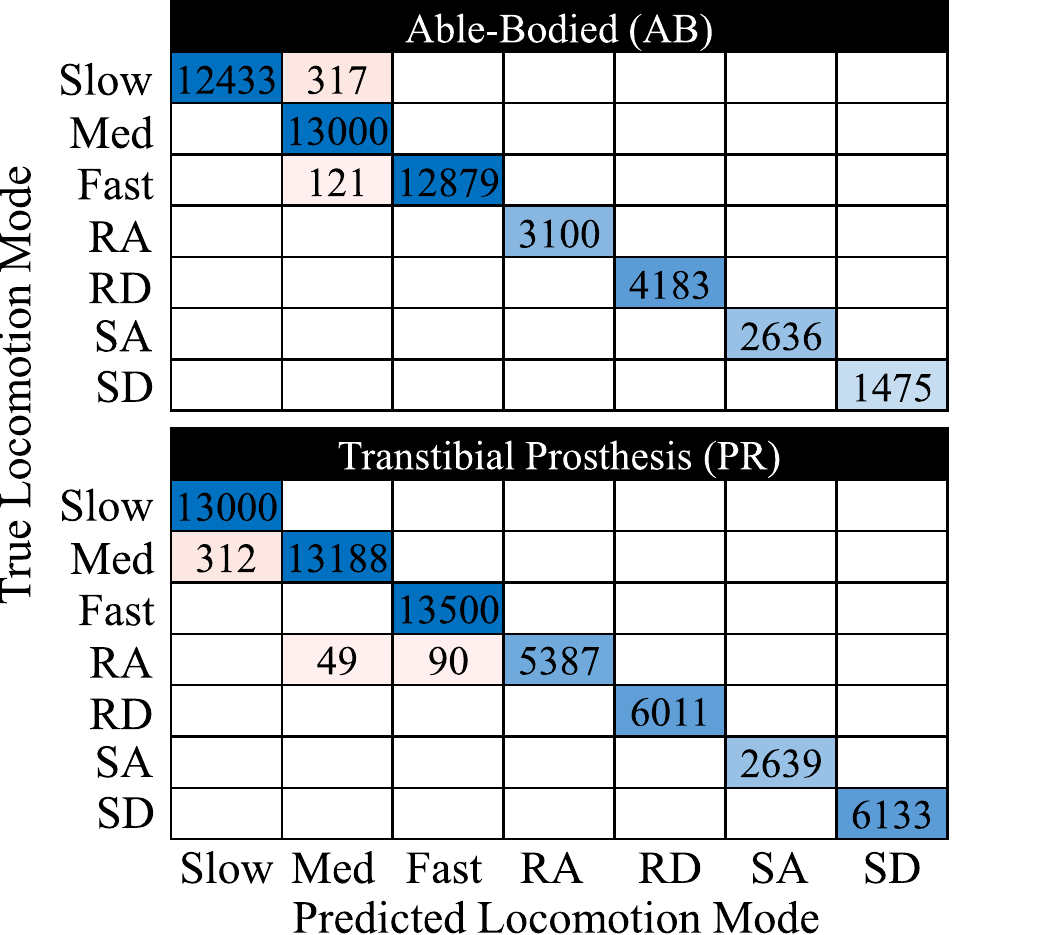}
    \caption{Confusion matrices summarizing the steady-state classification results for the able-bodied (top) and transtibial prosthesis (bottom) conditions. Slow, Med, Fast, RA, RD, SA, and SD correspond to steady state locomotion modes of 0.6, 0.8, and 1.0 m/s level walking, ramp ascent/descent, and stair ascent/descent, respectively.} 
	\label{fig:Confusion}
 \vspace{-0.5cm}
\end{figure}


The overall steady-state classification accuracy among locomotion modes 
is shown with the confusion matrices in Fig.~\ref{fig:Confusion} for both able-bodied (AB) and unilateral prosthesis (PR) conditions. In the AB condition, the overall classification accuracy was 99.1$\%$. Notably, misclassification errors in the AB condition only occurred during level walking. Only 2.5$\%$ of the predictions when walking at the slow speed (0.6 m/s) were for walking 0.2m/s faster (Med, 0.8 m/s), and 0.9$\%$ of the predictions when walking at the fast speed (1.0 m/s) were for walking 0.2m/s slower (Med, 0.8 m/s). When the true locomotion mode was steady-state level walking at 0.8 m/s, classification had an accuracy of 100\% (13,000 samples). Classification of all ramp walking (RA and RD) and stair climbing (SA and SD) was also 100$\%$ accurate.

\begin{figure}[t!]
    \centering
    \includegraphics[width = \linewidth, trim={0.2cm 15.3cm 0.4cm 0cm},clip]{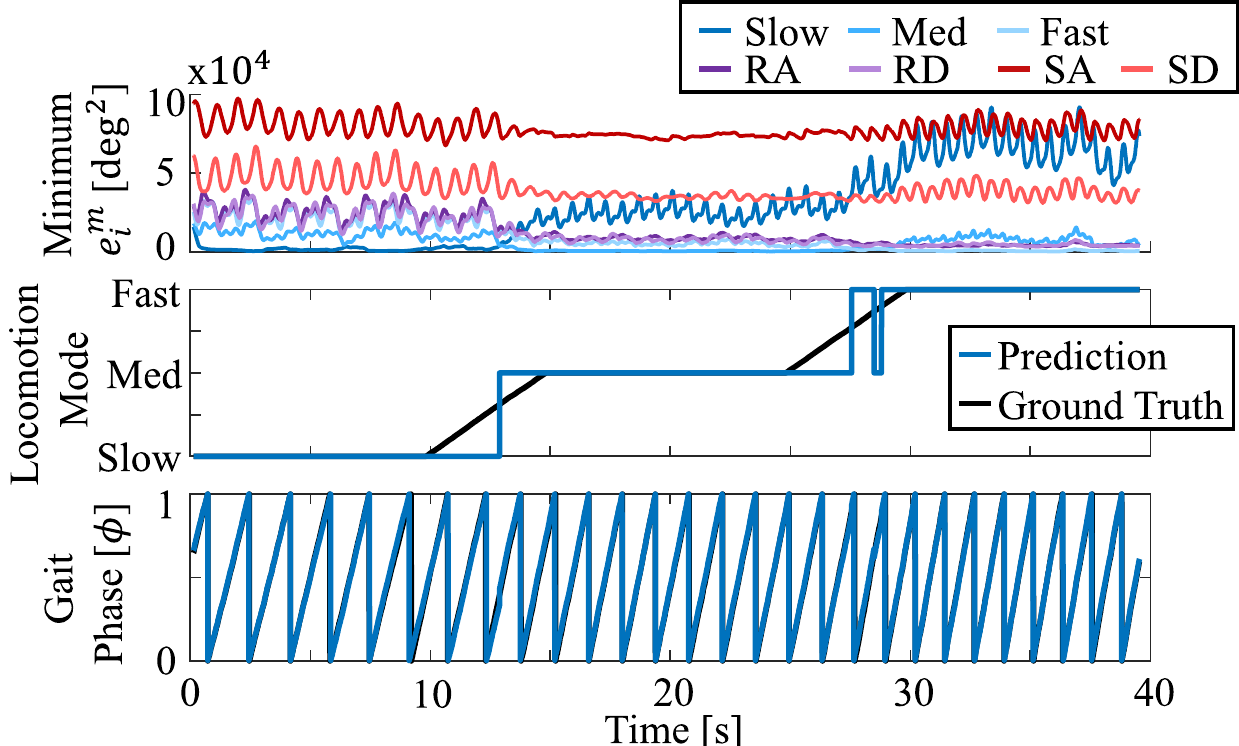}
    \caption{The lowest sum-of-squares error $e_i^m$ (top) is used to predict the locomotion mode (middle) and gait phase (bottom).} 
	\label{fig:Multispeed}
 \vspace{-0.5cm}
\end{figure}

The overall steady-state classification accuracy in the PR condition was 99.3$\%$. Similar to the AB condition, misclassification occurred with the speed identification during steady-state level walking, where 2.3$\%$ of the predictions when walking at 0.8 m/s were for walking at 0.6 m/s. During 5-degree ramp ascent, 0.8 and 1.0 m/s walking was incorrectly predicted 0.9$\%$ and 1.6$\%$ of the time, respectively. All fast walking, ramp descent, and stair ascent/descent for the PR condition was classified with 100$\%$ accuracy in these trials. In general, this high classification accuracy, paired with the relatively large number of locomotion modes considered ($M = 7$), many of which are very similar, is encouraging. A representative trial with the highly similar locomotion modes of slow, medium, and fast level walking speeds is shown in Fig.~\ref{fig:Multispeed}. 

\subsection{Gait Phase Estimation}

\begin{figure}[b!]
    \centering
    \includegraphics[width = 0.75\linewidth, trim={4cm 8.5cm 4cm 8.5cm},clip]{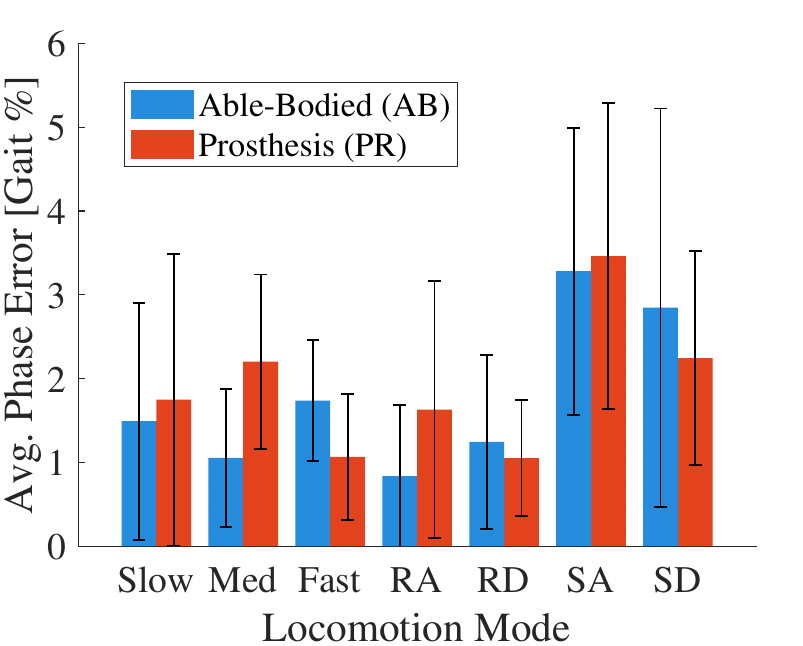}
    \caption{The average gait phase error ($\pm$ one standard deviation) for each locomotion mode is measured in gait percentage. The able-bodied (AB, blue) and prosthesis (PR, red) conditions demonstrated comparable gait phase estimation performance.} 
	\label{fig:PhaseBar}
 \vspace{-0.5cm}
\end{figure}

The average steady-state gait phase estimation error is shown in Fig.~\ref{fig:PhaseBar} for each locomotion mode and both AB and PR conditions. For this analysis, phase estimation was only considered when the locomotion mode was properly classified (99.1$\%$ of AB data and 99.3$\%$ of PR data). For level walking tasks, the phase error was generally below 1.8$\%$ in the AB condition and 2.3$\%$ in the PR condition on average. Ramp ascent was the locomotion mode with the lowest average phase error for the AB condition (0.8$\%$) while ramp descent yielded the lowest error in the PR condition (1.1$\%$). Both conditions had the largest average phase error during stair ascent (AB: 3.3$\%$, PR: 3.4$\%$), followed by stair descent (AB: 2.8$\%$, PR: 2.2$\%$).

During the instances of misclassification, gait phase was still estimated by the algorithm, but using a reference locomotion mode that was inconsistent with the true locomotion mode. Even during these misclassifications, the maximum gait phase error never exceeded 10$\%$ of the ground truth gait phase (maximum error observed of 7.6$\%$ and 9.1$\%$ for the AB and PR conditions, respectively). This further suggests that the few misclassifications that did occur ($< 1\%$) may not have resulted in critical failures when used for control.

\subsection{Computation Time}

The computation speed of the proposed method is crucial for online data processing on hardware. Table \ref{tab:Performance} provides benchmark results for the proposed method on two different platforms: a laptop (Apple MacBook Pro with M3 Pro chip) and a Raspberry Pi 4B (4GB RAM version), which is the default microcontroller of the target device (i.e., OSL prosthesis with Raspberry Pi). To fully demonstrate the computational efficiency of the proposed method, we implemented three versions using C++ \cite{SWGE2024} and conducted experiments over 10,000 time steps, each involving 7 activities, to compute the average computation time for each time step with new data.
The result in Table \ref{tab:Performance} shows that the naive method is not fast enough for a real-time processing on the target device since it can only process the new data under 110 Hz, which is slower than the sensing frequency (average 230 Hz). 

On the other hand, the proposed computation reduction approach, namely the sliding window algorithm, leads to a significant improvement in computation time: it is 310 or 1000 times faster compared to the naive method on the target device or MacBook. The variance in reduction between different machines stems from the CPU cache size: the reduction algorithm requires more memory ($\mathcal{O}(M\cdot N^2)$) compared to the naive method ($\mathcal{O}(M\cdot N)$). Consequently, the naive method can operate within the L2 cache of both machines, while the reduction approach exceeds the cache size of the Raspberry Pi. 
Regardless, with such low computation time, many more locomotion modes could be added to the analysis to enable a wider range of activities and to gain further insights. Overall, the proposed method incorporating the computation reduction algorithm can potentially handle 700 different activities with an updating frequency of up to 330 Hz on the OSL without any parallelization.


\begin{table}[t!]
\caption{Computation comparison for 10,000 data sample\\[-3ex]}
\vspace{-0.4cm}
\label{tab:Performance}
\begin{center}
\begin{tabular}{|c|c|c|}
\rowcolor[HTML]{333333} 
{\color[HTML]{FFFFFF} Method} & {\color[HTML]{FFFFFF} \begin{tabular}[c]{@{}c@{}}Ras-Pi 4B\\ Mean Time ($\mu$s)\end{tabular}} & {\color[HTML]{FFFFFF} \begin{tabular}[c]{@{}c@{}}MacBook M3 Pro\\ Mean Time ($\mu$s)\end{tabular}} \\ \hline
Naive Method                  & 9312.45 & 3023.34   \\ \hline
Naive Method with Eigen       & 1678.96 & 511.11    \\ \hline
Efficient Method              & \textbf{29.78} & \textbf{2.91} \\ \hline                           
\end{tabular} 
\end{center}
\vspace{-0.5cm}
\end{table}

\section{DISCUSSION}
\subsection{Interpretability}

Observing Fig.~\ref{fig:Multispeed}, this approach is transparently interpretable, and could be used to offer further insight in future implementations. For example, the sum-of-squares error between the measured joint angles and the reference joint angles is observed in real-time, and this could be converted to a root-mean-square value if desired. In doing so, not only could the gait phase be monitored, but the differences from expected trajectories could be used clinically to identify gait pathologies, or used in rehabilitation settings to provide real time feedback to users and therapists during gait training. These error values could also be used to compare the differences or similarities between the predicted locomotion mode and those modes that were not selected. For example, in Fig.~\ref{fig:Multispeed}, the medium and fast speeds are correctly classified, however, the thigh and shank kinematics suggest that the locomotion is also very similar to that of 5-degree ramp ascent and descent. This insight could aid in interpreting misclassifications and could also lead to including confidence levels in the predictions being made. While walking, if two different walking speeds are identified with similar error results, a low prediction confidence could accompany the locomotion mode classification, but with a very high confidence that the prediction is either one of the two speeds. 

\subsection{Limitations and Future Work}

With this method relying on historical data when determining the current locomotion mode and gait phase, the high classification accuracy observed is linked to steady-state locomotion. An inherent limitation of this approach is a delay when transitioning from one locomotion mode to another, unless the transition is gradual, as demonstrated in Fig.~\ref{fig:Multispeed}. 
Future work could investigate the incorporation of partial gait cycle kernels or locomotion transition trajectories such as in~\cite{cheng2023controlling}. High level supervision, such as a state-machine, could also ensure that only certain transitions are considered.

In this work, the classification and gait phase estimation make use of bilateral sensing, 
as in~\cite{lee2019classification, wu2019locomotion, zhang2022gait, weigand2022continuous, zhang2023real}. Many have used unilateral sensing~\cite{camargo2021machine, gao2020imu, bartlett2017phase, young2013training, huang2011continuous, varol2009multiclass, young2014intent, simon2016delaying, su2019cnn, feng2019strain, martinez2017simultaneous}, as bilateral sensing is often negatively associated with increased donning and doffing requirements, especially for unilateral prosthesis users who already experience high equipment demands. While it is generally undesirable to increase these demands, if it can be shown that the incorporation of additional IMUs (four total in this study) leads to high performance, the benefit may outweigh the cost. Extensive research taking place in the wearable sensors and wearable robotics fields~\cite{esfahani2018smart, ancans2021wearable} could also reduce this additional donning/doffing cost to be functionally equivalent to putting on an additional undergarment. Future work could investigate the performance of this algorithm when using unilateral sensing or single IMU sensing.

\section{CONCLUSION}
This paper presents a novel, interpretable, and computationally efficient algorithm to simultaneously predict locomotion mode and gait phase. This approach can be used in healthy gait analysis, asymmetrical gait, and towards the control of wearable robotics, as demonstrated by both able-bodied and transtibial prosthesis validation. The locomotion mode classification and gait phase estimation accuracy are high, especially when considering both the number of locomotion modes used ($M = 7$) and the similarity between these modes (three level walking modes with 0.2m/s speed difference). Due to the computational efficiency of this method, a much larger set of locomotion modes could be theoretically employed, to better accommodate the infinite set of locomotion modes pursued by humans in daily living (infinite ramp slopes, stair heights, and other activities). The interpretability of the presented algorithm also makes it ideal for better understanding misclassifications and improving quantifiable gait analysis.










\pagebreak

\bibliographystyle{IEEEtran}
\bibliography{References}

\begin{thebibliography}{10}
\providecommand{\url}[1]{#1}
\csname url@samestyle\endcsname
\providecommand{\newblock}{\relax}
\providecommand{\bibinfo}[2]{#2}
\providecommand{\BIBentrySTDinterwordspacing}{\spaceskip=0pt\relax}
\providecommand{\BIBentryALTinterwordstretchfactor}{4}
\providecommand{\BIBentryALTinterwordspacing}{\spaceskip=\fontdimen2\font plus
\BIBentryALTinterwordstretchfactor\fontdimen3\font minus \fontdimen4\font\relax}
\providecommand{\BIBforeignlanguage}[2]{{%
\expandafter\ifx\csname l@#1\endcsname\relax
\typeout{** WARNING: IEEEtran.bst: No hyphenation pattern has been}%
\typeout{** loaded for the language `#1'. Using the pattern for}%
\typeout{** the default language instead.}%
\else
\language=\csname l@#1\endcsname
\fi
#2}}
\providecommand{\BIBdecl}{\relax}
\BIBdecl

\bibitem{hoffman2019evaluating}
G.~Hoffman, ``Evaluating fluency in human--robot collaboration,'' \emph{IEEE Transactions on Human-Machine Systems}, vol.~49, no.~3, pp. 209--218, 2019.

\bibitem{gambon2020effects}
T.~M. Gambon, J.~P. Schmiedeler, and P.~M. Wensing, ``Effects of user intent changes on onboard sensor measurements during exoskeleton-assisted walking,'' \emph{IEEE Access}, vol.~8, pp. 224\,071--224\,082, 2020.

\bibitem{hutabarat2020quantitative}
Y.~Hutabarat, D.~Owaki, and M.~Hayashibe, ``Quantitative gait assessment with feature-rich diversity using two imu sensors,'' \emph{IEEE Transactions on Medical Robotics and Bionics}, vol.~2, no.~4, pp. 639--648, 2020.

\bibitem{caramia2018imu}
C.~Caramia, D.~Torricelli, M.~Schmid, A.~Munoz-Gonzalez, J.~Gonzalez-Vargas, F.~Grandas, and J.~L. Pons, ``Imu-based classification of parkinson's disease from gait: A sensitivity analysis on sensor location and feature selection,'' \emph{IEEE journal of biomedical and health informatics}, vol.~22, no.~6, pp. 1765--1774, 2018.

\bibitem{carcreff2018best}
L.~Carcreff, C.~N. Gerber, A.~Paraschiv-Ionescu, G.~De~Coulon, C.~J. Newman, S.~Armand, and K.~Aminian, ``What is the best configuration of wearable sensors to measure spatiotemporal gait parameters in children with cerebral palsy?'' \emph{Sensors}, vol.~18, no.~2, p. 394, 2018.

\bibitem{filli2018profiling}
L.~Filli, T.~Sutter, C.~S. Easthope, T.~Killeen, C.~Meyer, K.~Reuter, L.~L{\"o}rincz, M.~Bolliger, M.~Weller, A.~Curt \emph{et~al.}, ``Profiling walking dysfunction in multiple sclerosis: characterisation, classification and progression over time,'' \emph{Scientific reports}, vol.~8, no.~1, p. 4984, 2018.

\bibitem{ilias2017using}
T.~Ilias, B.~Filip, C.~Radu, N.~Dag, S.~Marina, and M.~Mevludin, ``Using measurements from wearable sensors for automatic scoring of parkinson's disease motor states: Results from 7 patients,'' in \emph{2017 39th Annual International Conference of the IEEE Engineering in Medicine and Biology Society (EMBC)}.\hskip 1em plus 0.5em minus 0.4em\relax IEEE, 2017, pp. 131--134.

\bibitem{del2015validation}
S.~Del~Din, A.~Godfrey, and L.~Rochester, ``Validation of an accelerometer to quantify a comprehensive battery of gait characteristics in healthy older adults and parkinson's disease: toward clinical and at home use,'' \emph{IEEE journal of biomedical and health informatics}, vol.~20, no.~3, pp. 838--847, 2015.

\bibitem{tucker2015control}
M.~R. Tucker, J.~Olivier, A.~Pagel, H.~Bleuler, M.~Bouri, O.~Lambercy, J.~d.~R. Mill{\'a}n, R.~Riener, H.~Vallery, and R.~Gassert, ``Control strategies for active lower extremity prosthetics and orthotics: a review,'' \emph{Journal of neuroengineering and rehabilitation}, vol.~12, pp. 1--30, 2015.

\bibitem{labarriere2020machine}
F.~Labarri{\`e}re, E.~Thomas, L.~Calistri, V.~Optasanu, M.~Gueugnon, P.~Ornetti, and D.~Laroche, ``Machine learning approaches for activity recognition and/or activity prediction in locomotion assistive devices—a systematic review,'' \emph{Sensors}, vol.~20, no.~21, p. 6345, 2020.

\bibitem{camargo2021machine}
J.~Camargo, W.~Flanagan, N.~Csomay-Shanklin, B.~Kanwar, and A.~Young, ``A machine learning strategy for locomotion classification and parameter estimation using fusion of wearable sensors,'' \emph{IEEE Transactions on Biomedical Engineering}, vol.~68, no.~5, pp. 1569--1578, 2021.

\bibitem{gao2020imu}
F.~Gao, G.~Liu, F.~Liang, and W.-H. Liao, ``Imu-based locomotion mode identification for transtibial prostheses, orthoses, and exoskeletons,'' \emph{IEEE Transactions on Neural Systems and Rehabilitation Engineering}, vol.~28, no.~6, pp. 1334--1343, 2020.

\bibitem{bartlett2017phase}
H.~L. Bartlett and M.~Goldfarb, ``A phase variable approach for imu-based locomotion activity recognition,'' \emph{IEEE transactions on biomedical engineering}, vol.~65, no.~6, pp. 1330--1338, 2017.

\bibitem{young2013training}
A.~J. Young, A.~M. Simon, and L.~J. Hargrove, ``A training method for locomotion mode prediction using powered lower limb prostheses,'' \emph{IEEE Transactions on Neural Systems and Rehabilitation Engineering}, vol.~22, no.~3, pp. 671--677, 2013.

\bibitem{chen2014locomotion}
B.~Chen, E.~Zheng, and Q.~Wang, ``A locomotion intent prediction system based on multi-sensor fusion,'' \emph{Sensors}, vol.~14, no.~7, pp. 12\,349--12\,369, 2014.

\bibitem{huang2011continuous}
H.~Huang, F.~Zhang, L.~J. Hargrove, Z.~Dou, D.~R. Rogers, and K.~B. Englehart, ``Continuous locomotion-mode identification for prosthetic legs based on neuromuscular--mechanical fusion,'' \emph{IEEE Transactions on Biomedical Engineering}, vol.~58, no.~10, pp. 2867--2875, 2011.

\bibitem{varol2009multiclass}
H.~A. Varol, F.~Sup, and M.~Goldfarb, ``Multiclass real-time intent recognition of a powered lower limb prosthesis,'' \emph{IEEE Transactions on Biomedical Engineering}, vol.~57, no.~3, pp. 542--551, 2009.

\bibitem{shin2021locomotion}
D.~Shin, S.~Lee, and S.~Hwang, ``Locomotion mode recognition algorithm based on gaussian mixture model using imu sensors,'' \emph{Sensors}, vol.~21, no.~8, p. 2785, 2021.

\bibitem{young2014intent}
A.~J. Young, A.~M. Simon, N.~P. Fey, and L.~J. Hargrove, ``Intent recognition in a powered lower limb prosthesis using time history information,'' \emph{Annals of biomedical engineering}, vol.~42, pp. 631--641, 2014.

\bibitem{simon2016delaying}
A.~M. Simon, K.~A. Ingraham, J.~A. Spanias, A.~J. Young, S.~B. Finucane, E.~G. Halsne, and L.~J. Hargrove, ``Delaying ambulation mode transition decisions improves accuracy of a flexible control system for powered knee-ankle prosthesis,'' \emph{IEEE Transactions on Neural Systems and Rehabilitation Engineering}, vol.~25, no.~8, pp. 1164--1171, 2016.

\bibitem{su2019cnn}
B.-Y. Su, J.~Wang, S.-Q. Liu, M.~Sheng, J.~Jiang, and K.~Xiang, ``A cnn-based method for intent recognition using inertial measurement units and intelligent lower limb prosthesis,'' \emph{IEEE Transactions on Neural Systems and Rehabilitation Engineering}, vol.~27, no.~5, pp. 1032--1042, 2019.

\bibitem{lee2019classification}
S.-S. Lee, S.~T. Choi, and S.-I. Choi, ``Classification of gait type based on deep learning using various sensors with smart insole,'' \emph{Sensors}, vol.~19, no.~8, p. 1757, 2019.

\bibitem{feng2019strain}
Y.~Feng, W.~Chen, and Q.~Wang, ``A strain gauge based locomotion mode recognition method using convolutional neural network,'' \emph{Advanced Robotics}, vol.~33, no.~5, pp. 254--263, 2019.

\bibitem{le2024transfer}
D.~Le, S.~Cheng, R.~D. Gregg, and M.~Ghaffari, ``Transfer learning for efficient intent prediction in lower-limb prosthetics: A strategy for limited datasets,'' \emph{IEEE Robotics and Automation Letters}, 2024.

\bibitem{sup2007design}
F.~Sup, A.~Bohara, and M.~Goldfarb, ``Design and control of a powered knee and ankle prosthesis,'' in \emph{Proceedings 2007 IEEE International Conference on Robotics and Automation}.\hskip 1em plus 0.5em minus 0.4em\relax IEEE, 2007, pp. 4134--4139.

\bibitem{posh2023finite}
R.~R. Posh, J.~P. Schmiedeler, and P.~M. Wensing, ``Finite-state impedance and direct myoelectric control for robotic ankle prostheses: Comparing their performance and exploring their combination,'' \emph{IEEE Transactions on Neural Systems and Rehabilitation Engineering}, vol.~31, pp. 2778--2788, 2023.

\bibitem{vu2020review}
H.~T.~T. Vu, D.~Dong, H.-L. Cao, T.~Verstraten, D.~Lefeber, B.~Vanderborght, and J.~Geeroms, ``A review of gait phase detection algorithms for lower limb prostheses,'' \emph{Sensors}, vol.~20, no.~14, p. 3972, 2020.

\bibitem{gregg2013experimental}
R.~D. Gregg, T.~Lenzi, N.~P. Fey, L.~J. Hargrove, and J.~W. Sensinger, ``Experimental effective shape control of a powered transfemoral prosthesis,'' \emph{IEEE Int Conf Rehabil Robot}, pp. 1--7, 2013.

\bibitem{gregg2013towards}
R.~D. Gregg and J.~W. Sensinger, ``Towards biomimetic virtual constraint control of a powered prosthetic leg,'' \emph{IEEE T Control Syst Technol}, vol.~22, no.~1, pp. 246--254, 2013.

\bibitem{villarreal2014survey}
D.~J. Villarreal and R.~D. Gregg, ``A survey of phase variable candidates of human locomotion,'' \emph{IEEE Int Conf Eng Med Biol Soc (EMBC)}, pp. 4017--4021, 2014.

\bibitem{cortino2023data}
R.~J. Cortino, T.~K. Best, and R.~D. Gregg, ``Data-driven phase-based control of a powered knee-ankle prosthesis for variable-incline stair ascent and descent,'' \emph{IEEE Transactions on Medical Robotics and Bionics}, 2023.

\bibitem{holgate2009novel}
M.~A. Holgate, T.~G. Sugar, and A.~W. Bohler, ``A novel control algorithm for wearable robotics using phase plane invariants,'' in \emph{IEEE Int Conf Robot Autom (ICRA)}, 2009, pp. 3845--3850.

\bibitem{posh2023calibration}
R.~R. Posh, J.~A. Tittle, J.~P. Schmiedeler, and P.~M. Wensing, ``Calibration of a tibia-based phase variable for control of robotic transtibial prostheses,'' in \emph{2023 IEEE/RSJ International Conference on Intelligent Robots and Systems (IROS)}.\hskip 1em plus 0.5em minus 0.4em\relax IEEE, 2023, pp. 1--6.

\bibitem{posh2024hybrid}
R.~R. Posh, J.~A. Tittle, D.~J. Kelly, J.~P. Schmiedeler, and P.~M. Wensing, ``Hybrid volitional control of a robotic transtibial prosthesis using a phase variable impedance controller,'' in \emph{2024 IEEE International Conference on Robotics and Automation (ICRA)}.\hskip 1em plus 0.5em minus 0.4em\relax IEEE, 2024, pp. 4555--4561.

\bibitem{kang2021real}
I.~Kang, D.~D. Molinaro, S.~Duggal, Y.~Chen, P.~Kunapuli, and A.~J. Young, ``Real-time gait phase estimation for robotic hip exoskeleton control during multimodal locomotion,'' \emph{IEEE robotics and automation letters}, vol.~6, no.~2, pp. 3491--3497, 2021.

\bibitem{xu2021noninvasive}
D.~Xu and Q.~Wang, ``Noninvasive human-prosthesis interfaces for locomotion intent recognition: A review,'' \emph{Cyborg and Bionic Systems}, 2021.

\bibitem{martinez2017simultaneous}
U.~Martinez-Hernandez, I.~Mahmood, and A.~A. Dehghani-Sanij, ``Simultaneous bayesian recognition of locomotion and gait phases with wearable sensors,'' \emph{IEEE Sensors Journal}, vol.~18, no.~3, pp. 1282--1290, 2017.

\bibitem{wu2019locomotion}
X.~Wu, Y.~Ma, X.~Yong, C.~Wang, Y.~He, and N.~Li, ``Locomotion mode identification and gait phase estimation for exoskeletons during continuous multilocomotion tasks,'' \emph{IEEE Transactions on Cognitive and Developmental Systems}, vol.~13, no.~1, pp. 45--56, 2019.

\bibitem{zhang2022gait}
X.~Zhang, H.~Zhang, J.~Hu, J.~Zheng, X.~Wang, J.~Deng, Z.~Wan, H.~Wang, and Y.~Wang, ``Gait pattern identification and phase estimation in continuous multilocomotion mode based on inertial measurement units,'' \emph{IEEE Sensors Journal}, vol.~22, no.~17, pp. 16\,952--16\,962, 2022.

\bibitem{weigand2022continuous}
F.~Weigand, A.~H{\"o}hl, J.~Zeiss, U.~Konigorski, and M.~Grimmer, ``Continuous locomotion mode recognition and gait phase estimation based on a shank-mounted imu with artificial neural networks,'' in \emph{2022 IEEE/RSJ International Conference on Intelligent Robots and Systems (IROS)}.\hskip 1em plus 0.5em minus 0.4em\relax IEEE, 2022, pp. 12\,744--12\,751.

\bibitem{zhang2023real}
X.~Zhang, E.~Tricomi, F.~Missiroli, N.~Lotti, and L.~Masia, ``Real-time assistive control via imu locomotion mode detection in a soft exosuit: An effective approach to enhance walking metabolic efficiency,'' \emph{IEEE/ASME Transactions on Mechatronics}, 2023.

\bibitem{zhang2014effects}
F.~Zhang, M.~Liu, and H.~Huang, ``Effects of locomotion mode recognition errors on volitional control of powered above-knee prostheses,'' \emph{IEEE Transactions on Neural Systems and Rehabilitation Engineering}, vol.~23, no.~1, pp. 64--72, 2014.

\bibitem{hafner2007evaluation}
B.~J. Hafner, L.~L. Willingham, N.~C. Buell, K.~J. Allyn, and D.~G. Smith, ``Evaluation of function, performance, and preference as transfemoral amputees transition from mechanical to microprocessor control of the prosthetic knee,'' \emph{Archives of physical medicine and rehabilitation}, vol.~88, no.~2, pp. 207--217, 2007.

\bibitem{das2019novel}
R.~Das, N.~Hooda, and N.~Kumar, ``A novel approach for real-time gait events detection using developed wireless foot sensor module,'' \emph{IEEE Sensors Letters}, vol.~3, no.~6, pp. 1--4, 2019.

\bibitem{SWGE2024}
\BIBentryALTinterwordspacing
R.~R. Posh, S.~Li, and P.~M. Wensing, ``{SlidingWindowGaitEstimation - Example code},'' 9 2024. [Online]. Available: \url{https://github.com/ROAM-Lab-ND/SlidingWindowGaitEstimation}
\BIBentrySTDinterwordspacing

\bibitem{cheng2023controlling}
S.~Cheng, C.~A. Laubscher, and R.~D. Gregg, ``Controlling powered prosthesis kinematics over continuous transitions between walk and stair ascent,'' in \emph{2023 IEEE/RSJ International Conference on Intelligent Robots and Systems (IROS)}.\hskip 1em plus 0.5em minus 0.4em\relax IEEE, 2023, pp. 2108--2115.

\bibitem{esfahani2018smart}
M.~I.~M. Esfahani and M.~A. Nussbaum, ``A “smart” undershirt for tracking upper body motions: Task classification and angle estimation,'' \emph{IEEE sensors Journal}, vol.~18, no.~18, pp. 7650--7658, 2018.

\bibitem{ancans2021wearable}
A.~Ancans, M.~Greitans, R.~Cacurs, B.~Banga, and A.~Rozentals, ``Wearable sensor clothing for body movement measurement during physical activities in healthcare,'' \emph{Sensors}, vol.~21, no.~6, p. 2068, 2021.

\end{thebibliography}

\end{document}